\documentclass[conference]{IEEEtran}
\IEEEoverridecommandlockouts
\usepackage{amsmath,amssymb,amsfonts}
\usepackage{algorithmic}
\usepackage{graphicx}
\usepackage{textcomp}
\usepackage{xcolor}
\usepackage{booktabs}
\usepackage[
    backend=biber,
    style=science,
]{biblatex}
\providecommand{\keywords}[1]
{
  \small	
  \textbf{\textit{Keywords---}} #1
}

\begin{document}

\title{Automatic Extraction of Medication Names in Tweets as Named Entity Recognition\\
}

\author{Carol Anderson\textsuperscript{\textsection}, Bo Liu\textsuperscript{\textsection}, Anas Abidin\textsuperscript{\textsection}, Hoo-Chang Shin\textsuperscript{\textsection}, Virginia Adams\textsuperscript{\textsection}\\
NVIDIA / Santa Clara, California, USA\\
\texttt{\{carola;boli;aabidin;hshin;vadams\}@nvidia.com}
}

\maketitle

\begingroup\renewcommand\thefootnote{\textsection}
\footnotetext{authors contributed equally.}
\endgroup

\begin{abstract}
Social media posts contain potentially valuable information about medical conditions and health-related behavior. Biocreative VII Task 3 focuses on mining this information by recognizing mentions of medications and dietary supplements in tweets. We approach this task by fine tuning multiple BERT-style language models to perform token-level classification, and combining them into ensembles to generate final predictions. Our best system consists of five Megatron-BERT-345M models and achieves a strict F1 score of 0.764 on unseen test data.
\end{abstract}

\vspace{.3cm}


\begin{table*}[h]
    \centering
    \caption{Performance of models trained on the training set and evaluated on the development set. The ensemble listed here is the first ensemble described in \ref{ensembles}.}
    \begin{tabular}{@{}llrrrrrr@{}}
    \toprule
    && \multicolumn{3}{c}{Overlap} & \multicolumn{3}{c}{Strict} \\ \cmidrule(r){3-5} \cmidrule(l){6-8}
    Model & Vocabulary & Precision & Recall & F1 & Precision & Recall & F1 \\ 
    \midrule
    Megatron-BERT-345M & BERT large uncased & 0.88 & 0.85 & 0.86 & 0.85 & 0.82 & 0.84 \\
    Megatron-BERT-345M & BERT large cased & 0.81 & 0.84 & 0.83 & 0.77 & 0.79 & 0.77 \\
    BioMegatron-BERT-345M & BERT large uncased & 0.91 & 0.83 & 0.87 & 0.82 & 0.77 & 0.79 \\
    BioMegatron-BERT-345M & BERT large cased & 0.88 & 0.74 & 0.80 & 0.84 & 0.71 & 0.77 \\
    RoBERTa large & RoBERTa large & 0.85 & 0.82 & 0.83 & 0.78 & 0.77 & 0.78 \\
    BERTweet large & RoBERTa large & 0.81 & 0.86 & 0.83 & 0.78 & 0.83 & 0.81 \\
    \textbf{Ensemble of five different models} && \textbf{0.92} & \textbf{0.86} & \textbf{0.89} & \textbf{0.90} & \textbf{0.84} & \textbf{0.87} \\
    \bottomrule
    \end{tabular}
    \vspace{12pt}

    \label{tab:single_models}
\end{table*}

\begin{table}[h]
    \centering
    \caption{Performance of ensembles on the test set.}
    \resizebox{1\columnwidth}{!}{
    \begin{tabular}{@{}lrrrrrr@{}}
    \toprule
    & \multicolumn{3}{c}{Overlap} & \multicolumn{3}{c}{Strict} \\ \cmidrule(lr){2-4} \cmidrule(lr){5-7}
    Ensemble & Precision & Recall & F1 & Precision & Recall & F1 \\ 
    \midrule
    Five different models & 0.847 & 0.714 & 0.775 & 0.823 & 0.694 &  0.753  \\
    \textbf{Megatron-BERT-345M uncased} & \textbf{0.835} & \textbf{0.755} & \textbf{0.793} & \textbf{0.805} & \textbf{0.728} & \textbf{0.764} \\
    \bottomrule
    \end{tabular}}
    \vspace{12pt}
    \label{tab:ensembles}
\end{table}

\begin{table}
    \centering
    \caption{Hyperparameters}
    \resizebox{.8\columnwidth}{!}{
    \begin{tabular}{@{}lrr@{}}
    \toprule
    Model & Learning rate & Max epochs \\
    \midrule
    Megatron-BERT-345M & 5 x $10^{-5}$ & 30 \\
    BioMegatron-BERT-345M & 8 x $10^{-5}$ & 30 \\
    RoBERTa large & 8 x $10^{-6}$ & 40 \\
    BERTweet large & 8 x $10^{-6}$ & 40 \\
    \end{tabular}}
    \label{tab:hyperparameters}
\end{table}

\section{Introduction}
Posts on social networks represent an enormous source of potentially useful health-related information. Twitter users currently generate an estimated 500 million tweets per day\footnote{retrived from https://www.internetlivestats.com/twitter-statistics/, October 8, 2021
}. Studies have shown that tweets  can be used to monitor various health-related phenomena, including infectious disease outbreaks~\cite{signorini2011use}, adverse drug events ~\cite{cocos2017pharmacovigilance, oconnor2014pharmacovigilance}, and drug abuse \cite{kazemi2017systematic}. However, extraction of information from tweets is particularly challenging due to several characteristics of the tweet format. First, because tweets are short (limited to 140 characters), it can be difficult to unambiguously identify the topics or entities mentioned. Second, tweets are extremely noisy, often containing abbreviations, misspellings, emojis, hashtags, and urls. Third, tweets are lexically and syntactically quite different from the text typically used to pretrain the language models, such as BERT \cite{devlin2018bert}, that are the basis of current state-of-the-art information extraction  methods. Fourth, any particular entity type is only found in a small fraction of tweets, meaning that even a large collection of labeled tweets may contain only a few examples of any entity type.

In Biocreative VII Task 3, we are asked to extract mentions of medications or dietary supplements from tweets by pregnant users. The training and development sets together contain 127,125 tweets, of which 311 contain at least one mention of a medication or supplement. We approach this task purely as a token-level classification problem.  We use no handcrafted features other than a minor customization of the tokenizer used to preprocess tweets. We experiment with various BERT-style models, finding that the best performance is obtained with the Megatron-BERT-345M model ~\cite{shoeybi2019megatron}. Finally, we boost performance by using multiple models together as ensembles \cite{Hinton2015DistillingTK}. 

\section{Methods}

\subsection{Preprocessing} \label{preprocessing}

In the data provided for the challenge, entity labels were given in the form of character indices. We converted these into token labels by tokenizing the tweets with the standard spaCy English tokenizer, and assigning labels of \texttt{B-DRUG} or \texttt{I-DRUG} to tokens that were the first or non-first tokens, respectively, in a labeled entity. Based on analysis of errors during initial experiments, we added a short list of custom tokens to the spaCy tokenizer as infixes and prefixes, which ensured they would be split apart from surrounding text and treated as tokens. For example, we found that the medication \texttt{Zofran} appeared multiple times in the challenge data and was sometimes embedded within a larger token, as in the hashtag \texttt{\#LifeWithAZofranPump}. Since partial tokens cannot be tagged as entities, Zofran would be missed in the preceding example unless the hashtag was split apart.
The custom token list we used was [\texttt{zofran}, \texttt{Zofran}, \texttt{Concerta}, \texttt{shots}, \texttt{nitrous}, \texttt{$\backslash$U000feb14}, and \texttt{/}].  

Each of the models we used required an additional tokenization step prior to training. Megatron and BioMegatron models use the WordPiece tokenizer ~\cite{sennrich2015neural}, while RoBERTa and BERTweet use a byte-level version of byte-pair encoding \cite{Radford2019LanguageMA}. In each case, the tokens produced by the spaCy tokenizer were further split into subtokens using the relevant tokenizer prior to training.

The authors of BERTweet \cite{nguyen2020bertweet} reported that they performed additional preprocessing steps before pretraining on tweets. These steps included using the NLTK TweetTokenizer, converting user mentions and urls into the special tokens \texttt{@USER} and \texttt{HTTPURL}, respectively, and converting emojis into text strings. Although we fine tuned BERTweet, we did not perform any of these additional steps prior to fine tuning; instead, due to time constraints, we used the same spaCy-tokenized data for all the models we fine tuned.

\subsection{Models} \label{models}
We approached the extraction of medication mentions as a token-level classification task, as is common practice. We used three token labels: \texttt{B-DRUG}, \texttt{I-DRUG}, and \texttt{O}. We used the NeMo \footnote{https://github.com/NVIDIA/NeMo} code base for fine tuning and inference. 

We first experimented with several different BERT-style models that differ mainly in their pretraining data or pretraining methods. These models included Megatron-BERT-345M (a 345-million parameter model pretrained on general domain text) \cite{shoeybi2019megatron}, BioMegatron-BERT-345M (the same architecture as Megatron-BERT-345M, but pretrained on text from PubMed) \cite{shin2020biomegatron}, RoBERTa large (pretrained on general domain text) \cite{liu2019roberta}, and BERTweet large (the same architecture as RoBERTa large, but pretrained on tweets) ~\cite{nguyen2020bertweet}. For all models, we used a classification head consisting of a single fully-connected layer with a dropout level of 0.5. We trained the models using a batch size of 64 on eight V100 GPUs, using the adam optimizer and the learning rates shown in Table \ref{tab:hyperparameters}. We applied warmup annealing with a warmup ratio of 0.1 to the learning rate. We trained for the maximum number of epochs shown in Table \ref{tab:hyperparameters} and saved the checkpoint with the highest token-level F1 score on the development set. The numbers reported in Table \ref{tab:single_models} were calculated using the evaluation script provided by the challenge organizers. The test set metrics in Table \ref{tab:ensembles} are the official results provided by the challenge organizer.

\subsection{Ensembles} \label{ensembles}
For our submissions, we created two different ensembles. The first ensemble consisted of five different models: Megatron-BERT-345M-uncased, Megatron-BERT-345M-cased, BioMegatron-BERT-345M-uncased, RoBERTa large, and BERTweet large. Each model was trained on the training set, and we used the checkpoint that performed best based on the development set. To generate final token labels, we calculated a weighted average of the class probabilities produced by each model for each token, using the following weights: BioMegatron-BERT-345M-uncased: 1, Megatron-BERT-345M-uncased,: 2, Megatron-BERT-345M-cased: 1.2, RoBERTa large: 0.4, BERTweet large: 1.4. We chose these weights through a random search constrained to the range between 0 and 2 (inclusive) in increments of 0.1. We initially also included a BioMegatron-BERT-345M-cased model in the ensemble, but we found that a weight of zero for that model gave the best performance on the development set, so we excluded it from the final ensemble.

The second ensemble consisted of five Megatron-BERT-345M models trained using the “out-of-fold” method. In this approach, we combined the training and development sets and then divided them randomly into five subsets. To train each model, we used four of these subsets as the training set and held out the fifth as a validation set. We used the checkpoint that performed best on the held-out set in each of the five runs in our final ensemble. At inference time, we calculated the mean of the token class probabilities from each of the five models and chose the token class with the highest probability.

\section{Results and Conclusions}
As shown in Table \ref{tab:single_models}, Megatron-BERT-345M-uncased gave the highest F1 scores of any single model when trained on the training set and evaluated on the development set. This ran counter to our expectation that BioMegatron-BERT would be better able to detect medications, given its pretraining on biomedical literature. A possible explanation for the poorer performance of BioMegatron-BERT compared to Megatron-BERT could be that tweets are linguistically more similar to the general domain text used to pretrain Megatron-BERT than to biomedical literature. In this regard, it is interesting to compare the peformance of the RoBERTa and BERTweet models, which share the same architecture but differ in their pretraining data. BERTweet modestly outperforms RoBERTa in the strict evaluation metrics. In terms of strict F1 scores, BERTweet was the second-best performing model after Megatron-BERT-345M-uncased. The performance of BERTweet could possibly be improved if the same preprocessing steps used prior to BERTweet pretraining (described in \ref{preprocessing} above) were also applied before fine tuning.

The performance of our two ensembles on the test set is shown in Table \ref{tab:ensembles}. The ensemble consisting entirely of Megatron-BERT-345M models outperformed the ensemble of different models, which again was contrary to our expectations. We expected that the ensemble of different models trained on the same data would outperform the ensemble consisting of a single model type trained on different tranches of data. Our results suggest that performance on this task is limited more by the training data than by model architecture. We also note that the performance of the ensemble of different models on the test set was much lower than its performance on the development set. This could be partially explained by the fact that we effectively overfitted to the development set by using it to choose the "best" checkpoint from each training run. However, the gap is so large that we suspect it indicates a significant difference in the distribution of text found in the development and test sets. 

\printbibliography

@article{sennrich2015neural,
  title={Neural machine translation of rare words with subword units},
  author={Sennrich, Rico and Haddow, Barry and Birch, Alexandra},
  journal={arXiv preprint arXiv:1508.07909},
  year={2015}
}

@article{devlin2018bert,
  title={Bert: Pre-training of deep bidirectional transformers for language understanding},
  author={Devlin, Jacob and Chang, Ming-Wei and Lee, Kenton and Toutanova, Kristina},
  journal={arXiv preprint arXiv:1810.04805},
  year={2018}
}

@article{shoeybi2019megatron,
  title={Megatron-lm: Training multi-billion parameter language models using model parallelism},
  author={Shoeybi, Mohammad and Patwary, Mostofa and Puri, Raul and LeGresley, Patrick and Casper, Jared and Catanzaro, Bryan},
  journal={arXiv preprint arXiv:1909.08053},
  year={2019}
}

@article{liu2019roberta,
  author    = {Yinhan Liu and
               Myle Ott and
               Naman Goyal and
               Jingfei Du and
               Mandar Joshi and
               Danqi Chen and
               Omer Levy and
               Mike Lewis and
               Luke Zettlemoyer and
               Veselin Stoyanov},
  title     = {RoBERTa: {A} Robustly Optimized {BERT} Pretraining Approach},
  journal   = {CoRR},
  volume    = {abs/1907.11692},
  year      = {2019},
  url       = {http://arxiv.org/abs/1907.11692},
  archivePrefix = {arXiv},
  eprint    = {1907.11692},
  bibsource = {dblp computer science bibliography, https://dblp.org}
}

@misc{shin2020biomegatron,
      title={BioMegatron: Larger Biomedical Domain Language Model},
      author={Hoo-Chang Shin and Yang Zhang and Evelina Bakhturina and Raul Puri and Mostofa Patwary and Mohammad Shoeybi and Raghav Mani},
      year={2020},
      eprint={2010.06060},
      archivePrefix={arXiv},
      primaryClass={cs.CL}
}

@inproceedings{nguyen2020bertweet,
title     = {{BERTweet: A pre-trained language model for English Tweets}},
author    = {Dat Quoc Nguyen and Thanh Vu and Anh Tuan Nguyen},
booktitle = {Proceedings of the 2020 Conference on Empirical Methods in Natural Language Processing: System Demonstrations},
pages     = {9--14},
year      = {2020}
}

@inproceedings{Radford2019LanguageMA,
  title={Language Models are Unsupervised Multitask Learners},
  author={Alec Radford and Jeff Wu and Rewon Child and David Luan and Dario Amodei and Ilya Sutskever},
  year={2019}
}

@article{cocos2017pharmacovigilance,
    author = {Cocos, Anne and Fiks, Alexander G and Masino, Aaron J},
    title = "{Deep learning for pharmacovigilance: recurrent neural network architectures for labeling adverse drug reactions in Twitter posts}",
    journal = {Journal of the American Medical Informatics Association},
    volume = {24},
    number = {4},
    pages = {813-821},
    year = {2017},
    month = {02},
    issn = {1067-5027},
    doi = {10.1093/jamia/ocw180},
    url = {https://doi.org/10.1093/jamia/ocw180},
    eprint = {https://academic.oup.com/jamia/article-pdf/24/4/813/34148877/ocw180.pdf},
}

@inproceedings{oconnor2014pharmacovigilance,
  title={Pharmacovigilance on twitter? Mining tweets for adverse drug reactions},
  author={O’Connor, Karen and Pimpalkhute, Pranoti and Nikfarjam, Azadeh and Ginn, Rachel and Smith, Karen L and Gonzalez, Graciela},
  booktitle={AMIA annual symposium proceedings},
  volume={2014},
  pages={924},
  year={2014},
  organization={American Medical Informatics Association}
}

@article{kazemi2017systematic,
  title={Systematic review of surveillance by social media platforms for illicit drug use},
  author={Kazemi, Donna M and Borsari, Brian and Levine, Maureen J and Dooley, Beau},
  journal={Journal of Public Health},
  volume={39},
  number={4},
  pages={763--776},
  year={2017},
  publisher={Oxford University Press}
}

@article{Hinton2015DistillingTK,
  title={Distilling the Knowledge in a Neural Network},
  author={Geoffrey E. Hinton and Oriol Vinyals and Jeffrey Dean},
  journal={ArXiv},
  year={2015},
  volume={abs/1503.02531}
}

@article{signorini2011use,
  title={The use of Twitter to track levels of disease activity and public concern in the US during the influenza A H1N1 pandemic},
  author={Signorini, Alessio and Segre, Alberto Maria and Polgreen, Philip M},
  journal={PloS one},
  volume={6},
  number={5},
  pages={e19467},
  year={2011},
  publisher={Public Library of Science San Francisco, USA}
}

% \bibliographystyle{IEEEtran}
% \bibliography{track_3}
% \bibitem{b1} G. Eason, B. Noble, and I. N. Sneddon, ``On certain integrals of Lipschitz-Hankel type involving products of Bessel functions,'' Phil. Trans. Roy. Soc. London, vol. A247, pp. 529--551, April 1955.
% \bibitem{b2} J. Clerk Maxwell, A Treatise on Electricity and Magnetism, 3rd ed., vol. 2. Oxford: Clarendon, 1892, pp.68--73.
% \bibitem{b3} I. S. Jacobs and C. P. Bean, ``Fine particles, thin films and exchange anisotropy,'' in Magnetism, vol. III, G. T. Rado and H. Suhl, Eds. New York: Academic, 1963, pp. 271--350.
% \bibitem{b4} K. Elissa, ``Title of paper if known,'' unpublished.
% \bibitem{b5} R. Nicole, ``Title of paper with only first word capitalized,'' J. Name Stand. Abbrev., in press.
% \bibitem{b6} Y. Yorozu, M. Hirano, K. Oka, and Y. Tagawa, ``Electron spectroscopy studies on magneto-optical media and plastic substrate interface,'' IEEE Transl. J. Magn. Japan, vol. 2, pp. 740--741, August 1987 [Digests 9th Annual Conf. Magnetics Japan, p. 301, 1982].
% \bibitem{b7} M. Young, The Technical Writer's Handbook. Mill Valley, CA: University Science, 1989.
% \end{thebibliography}
% \vspace{12pt}
\end{document}